\documentclass[10pt]{article}
\PassOptionsToPackage{dvipsnames,svgnames,x11names}{xcolor}  % named colors (BrickRed, ForestGreen, etc.)
\usepackage{colab}
\usepackage{lipsum}  % for demo text only — remove in real papers

\usepackage{mathrsfs}          % \mathscr script letters
\usepackage{dsfont}            % \mathds double-struck (e.g. \mathds{1} for indicator)
\usepackage{bbm}               % \mathbbm blackboard bold variants
\usepackage{upgreek}           % upright Greek letters (\upmu, \uppi, etc.)
\usepackage{siunitx}           % SI units typesetting (\SI{3.5}{\meter}, \num{1e6})

\usepackage{threeparttable}    % tables with footnotes
\usepackage{longtable}         % multi-page tables
\usepackage{rotating}          % sideways tables and figures
\usepackage{adjustbox}         % resize/clip boxes and figures
\usepackage{placeins}          % \FloatBarrier to control float placement
\usepackage{tabularx}           % flexible-width table columns (also in colab.sty)
\usepackage{colortbl}           % colored table cells
\usepackage{multirow}           % multi-row table cells
\usepackage{array}              % enhanced column definitions
\usepackage{makecell}           % multi-line cells, \thead, \makecell
\usepackage{diagbox}            % diagonal header cells (\diagbox{A}{B})
\usepackage{changepage}         % adjustwidth environment

\usepackage[normalem]{ulem}    % \uline, \sout (underline, strikethrough)
\usepackage{fancyvrb}           % fancy verbatim environments (\begin{Verbatim})
\usepackage{wasysym}            % extra symbols (\Square, \Circle, \smiley)
\usepackage{pgfplots}          % data-driven plots built on TikZ
\pgfplotsset{compat=1.18}
\usepackage{lineno}             % line numbers for peer review (\linenumbers to enable)
\usepackage{ifdraft}            % conditional commands for draft vs final mode

\usepackage{afterpage}          % execute command after current page ships out
\usepackage{pdflscape}          % landscape pages in a portrait document
\usepackage{bookmark}           % faster PDF bookmarks (improves hyperref bookmarks)

\usepackage{textcomp}           % extra text symbols (\texttimes, \textmu, etc.)
\usepackage{ragged2e}           % improved ragged/justified text (\RaggedRight)
\newcolumntype{Y}{>{\RaggedRight\arraybackslash}X}
\newcolumntype{P}[1]{>{\RaggedRight\arraybackslash}p{#1}}
\usepackage{calc}               % arithmetic in length/counter expressions

\usepackage{environ}            % capture environment body as a macro
\usepackage{xpatch}             % extend/patch commands and environments
\usepackage{ifthen}             % classic conditional commands

\title{Think Like a Pilot: Fine-Grained Long-Horizon UAV Navigation}

\author{%
  Xiangyi Zheng$^{1,*}$ \quad
  Xiangyu Wang$^{1,*}$ \quad
  Qinan Liao$^{1,*}$ \quad
  Zimu Tang$^{1}$ \quad
  Yue Liao$^{3}$ \quad
  \newline
  Dongyue Lyu$^{1}$ \quad
  Guodong Wang$^{2}$ \quad
  Junjie Liu$^{2}$ \quad
  Si Liu$^{1,\dagger}$ \\[6pt]
  {\normalsize $^{1}$Colab, Beihang University \quad $^{2}$Meituan \quad $^{3}$ National University of Singapore} \\[3pt]
  {\small $^{*}$Equal contribution \quad $^{\dagger}$Corresponding author: \texttt{liusi@buaa.edu.cn}}
}

\labname{Colab}
\institution{}
\colabdate{May 2026}
\paperurl{https://buaa-colalab.github.io/FLIGHT/}
\githuburl{https://github.com/buaa-colalab/FLIGHT}
\huggingfaceurl{https://huggingface.co/datasets/jujujulien/FLIGHT}
\colabrunningtitle{Think Like a Pilot: Fine-Grained Long-Horizon UAV Navigation}

\begin{document}

\maketitle

%% --- Abstract (Req 1: left-aligned title done in \maketitle) ---
%% --- Abstract (Req 2: hyphenation enabled, links after abstract) ---
%% --- Abstract (Req 3: light gray block, no frame, no title, no keywords) ---
\begin{colababstract}
Language-guided UAV agents must execute long-horizon semantic instructions while producing smooth, physically feasible continuous flight commands, yet existing Vision-Language Navigation (VLN) benchmarks typically use discrete or coarse actions and existing UAV Vision-Language-Action (VLA) tasks focus on short, atomic maneuvers. To address this gap in UAV task settings, we introduce \textbf{FLIGHT}, a \textbf{F}ine-grained \textbf{L}ong-horizon \textbf{I}nstruction-\textbf{G}uided benchmark for \textbf{H}ybrid UAV navigation and reasoning \textbf{T}asks, which combines multi-stage instructions with dense 6-DoF trajectory annotations across two dataset splits: Fine-grained VLN and Long-horizon Flow. To endow the UAV agent with the capability of real-time in-flight reasoning over task execution status and mission planning, while simultaneously accommodating high-frequency, real-time precise control, we further propose \textbf{FLIGHT VLA}, an asynchronous architecture that decouples a low-frequency Streaming Pilot Vision-Language Model (VLM) for task-state reasoning from a high-frequency diffusion action model for continuous control, supervised by explicit \textbf{Pilot Reasoning} texts that summarize the current flight state and anticipate the next subgoal. In closed-loop evaluation, FLIGHT VLA consistently surpasses representative VLN and VLA baselines on our FLIGHT benchmarks, achieving stronger multi-stage completion, subgoal adherence, and terminal control. Its trained Streaming Pilot Reasoning VLM further improves UAV video reasoning, validating the effectiveness of our design.

\vspace{0.4cm}
\colablinks
\end{colababstract}

\vspace{0.4cm}

%% ============================================================
%%  1. Introduction
%% ============================================================

\section{Introduction}

Unmanned Aerial Vehicles (UAVs) are currently undergoing a profound paradigm shift, evolving from specialized flight platforms into autonomous agents capable of executing complex tasks in open-ended environments. This evolution expands their utility across diverse scenarios, including logistics delivery, disaster rescue, and urban inspection. Moving away from traditional operations that rely on rigid waypoint planning or manual remote control, language-guided UAV control empowers users to directly specify mission objectives via natural language. Consequently, the UAV must autonomously comprehend commands, perceive its surroundings, and execute continuous flight maneuvers. This interactive paradigm holds great promise for lowering the barrier to UAV deployment and accelerating their transition from passive flight tools to embodied agents capable of autonomous aerial missions. 
% Achieving this ambitious goal, however, requires the underlying models to simultaneously master high-level semantic understanding, long-horizon task planning, environmental perception, and robust, low-level continuous control.

\begin{figure}[htbp]
  \centering
  \includegraphics[width=0.95\linewidth]{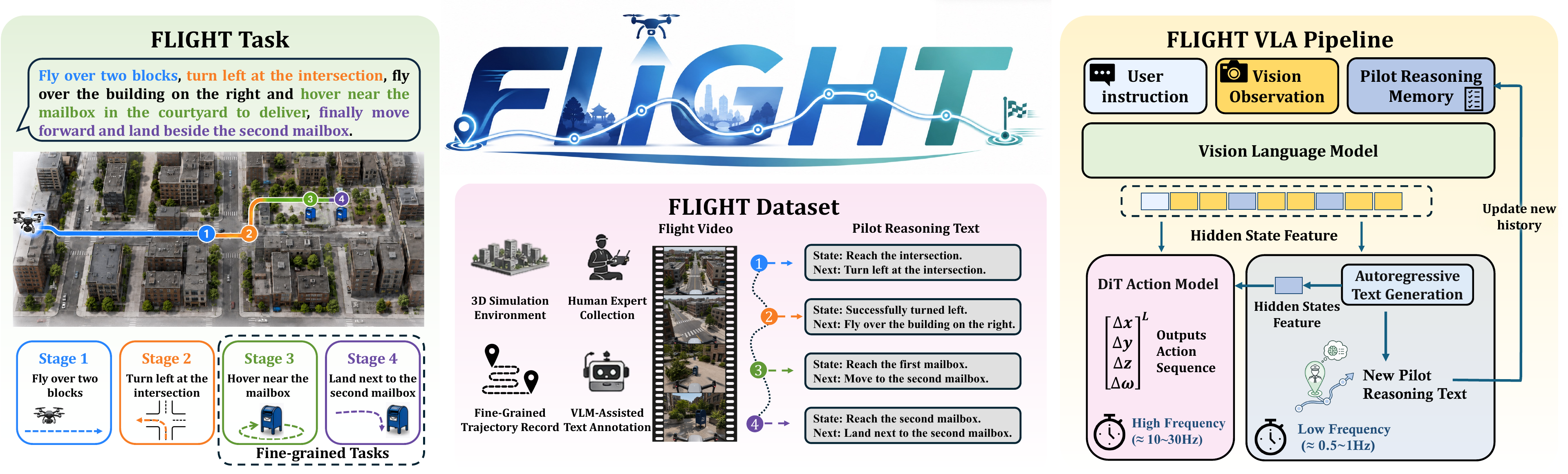}
  \caption{Overview of the FLIGHT benchmark and VLA framework. Left: FLIGHT targets the real-world scenario targeted where the UAV agent needs to possess both long-horizon navigation and fine-grained control capabilities. Middle: FLIGHT is constructed from high-quality flight videos collected in 3D simulation environments, paired with fine-grained trajectory records and language annotations; it further introduces the Pilot Reasoning paradigm to explicitly annotate the semantic logic behind navigation. Right: the proposed FLIGHT VLA framework asynchronously combines a streaming video VLM with a VLA action model and introduces Pilot Reasoning memory to bridge task-level decision reasoning and real-time efficient control.}
  \label{fig:pipeline}
\end{figure}

% Despite recent remarkable progress in Vision-Language Navigation (VLN) and Vision-Language-Action (VLA) models, existing benchmarks and task settings fail to adequately cover the full spectrum of capabilities required by an aerial embodied agent.

Recent advances in Vision-Language Navigation (VLN) ~\citep{liu_2023_AerialVLN,OpenFly,wang2024Travel,airstar} and Vision-Language-Action (VLA) ~\citep{nips2025_92cfa104,IndoorUAV,zhang2025embodied} have demonstrated remarkable potential in grounding language instructions to embodied perception and action, laying a solid foundation for the development of intelligent navigation systems. However, existing benchmarks and task settings do not fully encompass the comprehensive capability requirements of aerial embodied agents. On one hand, conventional VLN tasks primarily focus on long-range destination finding—essentially determining "where to go" based on language instructions. However, these tasks are predominantly built upon discrete action spaces or simplified kinetic models, which fail to characterize the continuous trajectories and 3D kinematic constraints inherent to real-world UAV flight. On the other hand, while recent studies have advanced fine-grained continuous trajectory control for UAVs, their instruction sets remain horizon-constrained and confined to isolated, atomic tasks—such as bypassing a single obstacle, approaching a local target, or executing short-range trajectory tracking. As aerial embodied agents, UAVs fundamentally demand a synergy of both paradigms: models must not only comprehend long-horizon, multi-stage semantic goals but also generate precise, smooth, and physically feasible continuous control commands at each execution stage. As illustrated in the left panel of Fig. \ref{fig:pipeline}, complex missions— such as "fly two blocks ahead, turn left at the intersection to hover and drop off a package near the mailbox in the courtyard, then proceed forward to finally land next to the second mailbox" can neither be reduced to traditional long-range navigation nor handled by short-term localized control alone.

This type of long-horizon, fine-grained missions exposes two core challenges. First, long-range language instructions demand that the model continuously maintain the task state throughout the flight, including tracking the current sub-goal, completed stages, next action intents, and their correspondences with environmental semantics. However, from an egocentric UAV perspective, the visual context undergoes drastic variations due to high-speed maneuvers and camera attitude changes. Consequently, models are highly susceptible to semantic drift, sub-goal forgetting, or erroneous task-completion assessments. Relying solely on end-to-end vision-to-action mapping often fails to maintain stable task-progress tracking over long-duration flights. Second, achieving robust semantic understanding and reasoning typically requires deploying heavy, high-level Vision-Language Models (VLMs); however, their inference frequencies are inherently incompatible with the real-time control demands of UAVs. Stable flight necessitates high-frequency, continuous, and low-latency action outputs. Directly utilizing VLMs to generate control commands frame-by-frame introduces severe computational latency, which inevitably degrades trajectory smoothness and compromises flight safety. Therefore, language-guided UAV control must fundamentally reconcile the temporal mismatch between long-horizon semantic planning and high-frequency continuous control.

To advance research on long-horizon, fine-grained UAV flight missions, we introduce FLIGHT - a novel benchmark for Fine-grained Long-horizon Instruction-Guided Hybrid UAV Tasks. FLIGHT challenges autonomous agents to execute closed-loop flights within semantically rich 3D environments, guided by multi-stage natural language instructions. To support this benchmark, we collected a large-scale dataset of continuous flight trajectories demonstrated by expert human pilots across diverse, semantically rich environments. Unlike traditional navigation models that rely on sparse and simplified waypoints, our approach captures dense, fine-grained control points. This ensures that the generated trajectories maintain high fidelity and exhibit rich physical feasibility aligned with actual flight dynamics. Rather than merely evaluating whether an agent reaches the final destination, FLIGHT rigorously assesses its capability to correctly execute intermediate sub-goals over long horizons while maintaining stable and accurate flight behaviors at the continuous control level.
To address the cognitive challenges inherent to such long-horizon missions, we introduce the \textbf{Pilot Reasoning mechanism}. Leveraging large-scale VLM-assisted annotation coupled with human verification, we externalize the "decision-making chain" of human pilots. Specifically, at each critical junction, the mechanism explicitly outputs an analysis of the current flight state along with a proactive planning of the subsequent action. As illustrated in the middle of Fig. \ref{fig:pipeline}, these Pilot Reasoning Text examples demonstrate how this explicit reasoning signal acts as a "cognitive anchor," effectively bridging the gap between high-level mission objectives and low-level control commands.

Furthermore, we propose the FLIGHT VLA architecture, which asynchronously decouples a slow semantic module—dedicated to streaming video understanding and task-state updates—from a fast action module tailored for high-frequency continuous control. The slow module maintains long-horizon context and periodically updates the semantic planning state, while the fast module operates at a significantly higher frequency to generate fine-grained flight control commands based on the immediate visual input and the latest planning state. Through this dual-system design, our model satisfies the real-time requirements of closed-loop UAV control while fully preserving its capacity for long-range language comprehension.

We construct a comprehensive closed-loop evaluation suite for FLIGHT in simulation, adapting representative VLN models, including LAG~\citep{liu_2023_AerialVLN} and NaVid~\citep{zhang2024navid}, alongside VLA models such as OpenVLA~\citep{kim24openvla} and MemoryVLA~\citep{shi2025memoryvla}, as baselines. Empirical results show that existing methods struggle with multi-stage UAV missions, mainly due to limited long-horizon state maintenance and continuous control capabilities. In contrast, FLIGHT VLA integrates Streaming VLM based on Pilot Reasoning chains of thought paradigm with an asynchronous, high-frequency real-time action model, achieving substantial performance gains across all evaluation metrics. These findings highlight the importance of explicit intermediate reasoning and asynchronous control for language-guided UAV tasks.

Our contributions are summarized as follows.
\begin{itemize}
  \item We propose \textbf{FLIGHT}, a long-horizon fine-grained hybrid-control paradigm and benchmark for UAV navigation, featuring semantically rich multi-stage trajectories and \textbf{Pilot Reasoning} supervision to bridge high-level navigation and low-level continuous control.
  \item We introduce \textbf{FLIGHT VLA}, an asynchronous fast–slow architecture that separates streaming semantic reasoning from real-time closed-loop control, enabling long-context understanding and stable continuous flight.
  \item We provide a systematic closed-loop evaluation of representative VLN and VLA baselines, demonstrating the challenges of FLIGHT and the effectiveness of Pilot Reasoning and FLIGHT VLA in long-horizon UAV control.
\end{itemize}

%% ============================================================
%%  2. Related Work
%% ============================================================
\section{Related Work}

\subsection{Language-Guided UAV Tasks}
Existing benchmarks for language-guided UAV tasks have primarily focused on vision-language navigation (VLN), where an agent is required to interpret high-level natural language instructions, plan accordingly, and complete navigation missions.
Aerial VLN~\citep{liu_2023_AerialVLN} represents flight trajectories as discrete action sequences for long-range navigation specified by language instructions.
Travel~\citep{wang2024Travel} introduces a waypoint-level autonomous UAV navigation task assisted by external instructions.
OpenFly~\citep{OpenFly} further provides a comprehensive cross-scene UAV VLN platform, covering 18 scenes and 100K trajectories, thereby substantially broadening the scope of UAV navigation in general-purpose environments.

Beyond the VLN paradigm, several studies have explored UAV motion control from the perspective of vision-language-action (VLA) tasks.
UAV-Flow~\citep{nips2025_92cfa104} first formulates language-guided fine-grained UAV motion control under the VLA paradigm through the Flow task (Flying-on-a-word), constructing a short-instruction dataset for common UAV maneuvers such as circling, traversing, and landing, and demonstrating the feasibility of VLA architectures for local fine-grained control.
Indoor UAV~\citep{IndoorUAV} builds an indoor UAV navigation dataset and further constructs IndoorUAV-VLA by selecting keyframes and regenerating concise instructions, thereby decomposing long trajectories into multiple shorter sub-trajectories.
Nevertheless, existing benchmarks do not effectively unify the long-horizon nature of navigation with the fine-grained control requirements of VLA tasks.
In contrast, our proposed FLIGHT benchmark introduces long-horizon VLA tasks that compose multiple fine-grained instructions (\textbf{Long-horizon Flow}) and incorporates fine-grained motion control into conventional navigation (\textbf{Fine-grained VLN}), thereby extending the frontier of language-guided UAV tasks.

\subsection{VLA with Reasoning}

In recent years, incorporating explicit reasoning processes into VLA models has emerged as an important paradigm for improving decision making in complex tasks.
These methods typically introduce intermediate reasoning representations in language, vision, or action spaces before action generation, thereby enhancing task decomposition, long-horizon planning, and interpretability.
For example, ECoT~\citep{Zawalski24-ecot}, CoT-VLA~\citep{zhao2025cotvla}, dVLA~\citep{wen2025dvladiffusionvisionlanguageactionmodel}, and ACoT-VLA~\citep{zhong2026acot} approach this problem from the perspectives of embodied chain-of-thought reasoning, visual subgoal prediction, multimodal CoT, and action-intention reasoning, respectively.
By enabling VLA models to generate intermediate reasoning processes related to both the environment state and task objective before execution, these methods improve generalization and decision stability in complex manipulation tasks.

Another line of work focuses more directly on high-level semantic understanding and task-structure modeling.
Methods such as $\pi_{0.5}$~\citep{intelligence2025pi05visionlanguageactionmodelopenworld} and ChatVLA-2~\citep{zhou2025visionlanguageactionmodelopenworldembodied} preserve or enhance the open-world understanding capability of pretrained VLMs, while introducing high-level semantic prediction or task decomposition mechanisms to better handle multi-stage robotic control tasks in open environments.
Meanwhile, to address long-horizon interaction and history-dependent decision making, MEM~\citep{torne2026memmultiscaleembodiedmemory}, VLingNav~\citep{wang2026vlingnav}, and LongNav-R1~\citep{hu2026longnavr1horizonadaptivemultiturnrl} explore mechanisms such as linguistic memory, adaptive reasoning, and reinforcement learning to mitigate context forgetting and model cross-step dependencies in long-horizon tasks.
ThinkAct~\citep{huang2025thinkact} and DeepThinkVLA~\citep{yin2025deepthinkvla} further adopt a ``think-before-act'' system perspective, optimizing the consistency between reasoning and action execution through reinforcement learning or causal alignment, thereby improving performance in few-shot adaptation, long-horizon planning, and autonomous error correction.
% Nevertheless, existing methods still face challenges in long-horizon navigation tasks, including insufficient task understanding, difficulty in managing cross-step context, and limited utilization of streaming video information.
Nevertheless, existing VLA approaches still exhibit notable limitations in long-horizon task comprehension and historical context modeling when deployed in scenarios such as drone navigation, where task execution spans extended durations and involves substantial viewpoint variations. Reasoning-based paradigms also face inherent challenges in balancing inference capability with the real-time requirements of action generation in highly dynamic drone navigation scenarios. To address these limitations, we propose a fast--slow VLA architecture that integrates streaming video reasoning, enhancing planning and contextual understanding in long-horizon scenarios.

%% ============================================================
%%  3. Method
%% ============================================================
% \begin{figure}[!t]
%   \centering
%   \includegraphics[width=0.95\linewidth,height=0.42\textheight,keepaspectratio]{images/task_compare.pdf}
%   \caption{Comparative analysis of our FLIGHT task setups against traditional VLN and fine-grained UAV flight tasks. (a) Expansion of our method over the UAV Flow task in terms of long-range trajectories. (b) Fine-grained characteristics of our VLN dataset regarding task instructions and action spaces.}
%   \label{fig:taskcompare}
% \end{figure}

% \section{FLIGHT}
\section{FLIGHT Benchmark and Dataset}

In this section, we present the FLIGHT benchmark, including its task formulation, UAV flight data collection methodology, and our proposed automated pipeline for generating instruction texts and Pilot Reasoning annotations.

\subsection{FLIGHT Task Definition}

FLIGHT comprises two types of UAV flight tasks: Long-horizon Flow and Fine-grained VLN. Long-horizon Flow inherits the Flying-on-a-Word task setting proposed in ~\citep{nips2025_92cfa104} and further extends it from single-stage atomic actions to multi-stage compositional action tasks. Fine-grained VLN, built upon the conventional VLN paradigm, introduces denser textual instruction descriptions and requires continuous fine-grained control. Fig. \ref{fig:taskcompare} shows the 

At time step $t$, the task input consists of the language instruction $I$, the recent visual observation window $V_{[t-W,t]}$ of length $W$, and the UAV sensor state $S_t$. The model is required to generate a fine-grained action sequence ($a_{t+1}$) to execute the task specified by the language instruction. Moreover, compared with the typical short-horizon task paradigm in VLA, relying solely on the observation input at the current time step is insufficient for completing long-sequence tasks. Therefore, in this task, the model typically needs to maintain historical information ($M_t$) and provide it to the decision-making process.

\begin{equation}
  \pi:(S_t,V_{[t-W,t]},I, {\color{Gray} M_t}) \mapsto a_{t+1},{\color{Gray} M_{t+1}}.
  \label{flight_task_def}
\end{equation}

\begin{figure}[!t]
  \centering
  \includegraphics[width=0.95\linewidth, keepaspectratio]{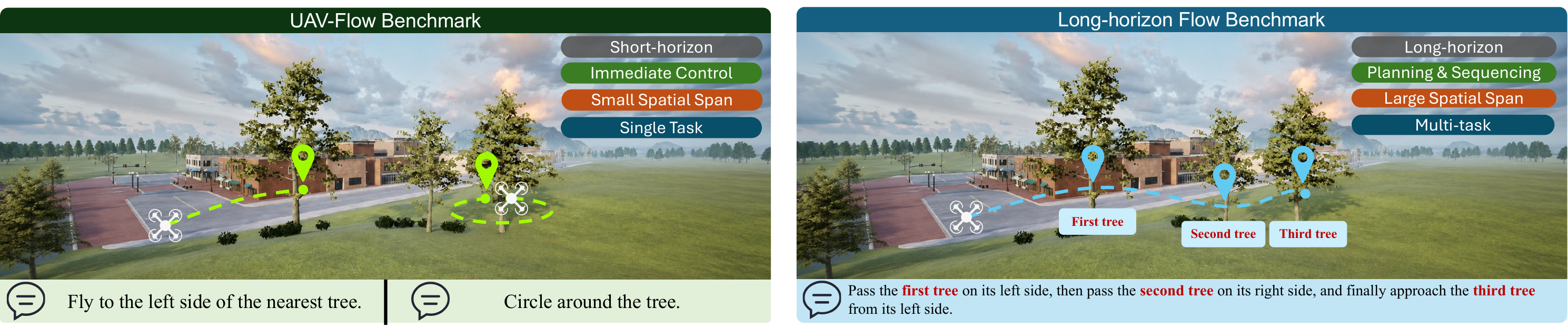}
  \caption{Comparison between UAV Flow task and ours FLIGHTLong-Horizon Flow task. Our task setting expands the UAV Flow task to support long-range trajectories.}
  \label{fig:task_compare1}
\end{figure}

In the \textbf{Long-horizon Flow} task, we inherit the rich set of atomic action types from UAV-Flow. The UAV is required to complete a variety of tasks according to natural language instructions, including translation, traversal, approaching, rotational orbiting, takeoff, and landing. As shown in Fig. \ref{fig:task_compare1}, our data expands the motion patterns in UAV-Flow from isolated short instructions to compositional hybrid actions. For example, instructions such as “traverse the first tree from the left, then traverse the second tree from the right,” “traverse the tree ahead and then land,” and “translate first and then orbit” define long-horizon motion paradigms that integrate multiple atomic actions. Appendix \ref{instruction_example} provides additional instruction examples.

The \textbf{Fine-grained} VLN task requires the UAV to perform navigation according to natural language instructions. To ensure that language instructions can be reliably aligned with spatial locations in real-world scenes, we provide explicit landmark descriptions at all task transition points, thereby improving the accuracy and consistency of language grounding.

\begin{figure}[!t]
  \centering
  \includegraphics[width=0.95\linewidth, keepaspectratio]{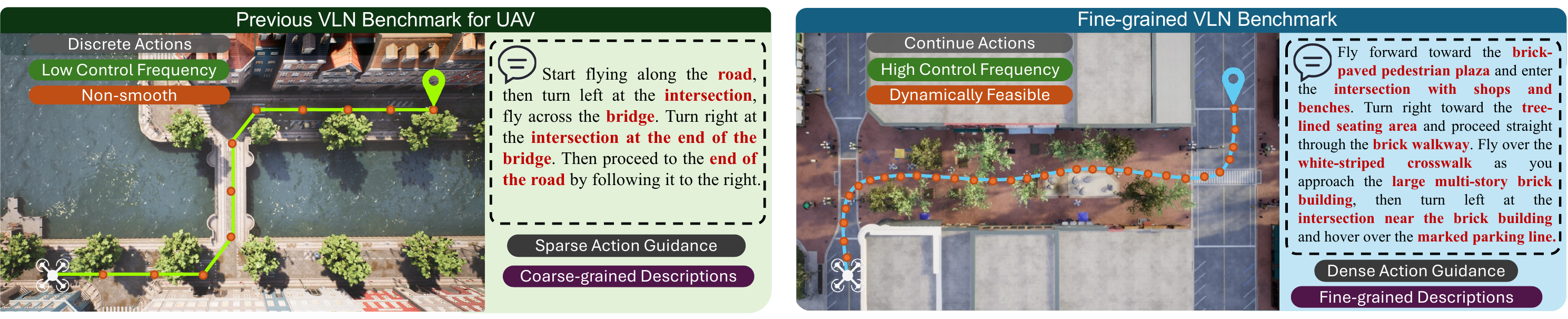}
  \caption{Comparison between previous Vision Language Navigation task and our FLIGHT-Fine-Grained VLN Task. FLIGHT Fine-Grained VLN exhibits fine-grained characteristics in both task instructions and action spaces.}
  \label{fig:task_compare2}
\end{figure}

In terms of instruction design, we follow the textual paradigm of natural language navigation instructions used in conventional VLN benchmarks, while preserving rich and diverse semantic expressions. As illustrated in Fig. ~\ref{fig:task_compare2}, compared with existing VLN datasets, our dataset enhances the fine-grained nature of the task from the following two aspects.

First, FLIGHT adopts a \textbf{finer-grained subtask decomposition} strategy. We divide the navigation process into a larger number of local subtasks, such that each unit of path length contains more decision-making actions. For example, in addition to instructions such as “go straight through the road,” we further incorporate local referential constraints such as “pass by the flower bed” and “move along the right side of the fountain.” This design increases the density of spatial grounding in the instructions, requiring the model to more frequently reason about the relationship between the local environment and the navigation goal. As shown in Table~\ref{tab:vln_granularity_stats} shows that Fine-Grained VLN provides significantly denser language grounding and finer-grained action supervision per unit trajectory length than Aerial VLN, thereby better supporting fine-grained subtask decomposition. Appendix \ref{instruction_example} provides additional examples illustrating the differences in instruction granularity between Fine-Grained VLN and previous VLN tasks.

\begin{table}[htbp]
  \centering
  \caption{Quantitative comparison between Aerial VLN and FLIGHT-fine-grained VLN. Despite having shorter average trajectories, FLIGHT-fine-grained VLN incorporates denser language grounding and finer action supervision.}
  \label{tab:vln_granularity_stats}
  \begin{threeparttable}
    \begin{tabularx}{\linewidth}{@{}>{\raggedright\arraybackslash}Xcc@{}}
      \toprule
      \textbf{Metric} & \textbf{Aerial VLN~\citep{liu_2023_AerialVLN}} & \textbf{FLIGHT-FG VLN} \\
      \midrule
      Average instruction length (words) & 83 & 71 \\
      Average trajectory length (m) & 661 & 154 \\
      Verbs per 100 m & 2.17 & 6.23 \\
      Nouns per 100 m & 3.25 & 12.26 \\
      Adjectives per 100 m & 0.96 & 6.55 \\
      Average action steps per 100 m\tnote{*} & 30.86 & 295.45 \\
      Control formulation & Discrete action classification & Continuous action sequence \\
      \bottomrule
    \end{tabularx}
    \begin{tablenotes}[flushleft]
      \footnotesize
      \item[*] In Aerial VLN, each action step denotes one discrete action choice. In FLIGHT-FG VLN, each action step denotes one element in the fine-grained continuous action sequence.
    \end{tablenotes}
  \end{threeparttable}
\end{table}

Second, \textbf{a continuous and fine-grained action space} is employed by FLIGHT. Unlike conventional VLN datasets, which primarily formulate navigation as discrete action classification, we record fine-grained action sequences from continuous UAV trajectories as the action space, thereby preserving the continuous control process in real-world flight. As a result, the task requires not only high-level path planning but also fine-grained motion control capabilities. The detailed design of the action space will be introduced in the following section.

Completing these two tasks requires a UAV agent to jointly perform hierarchical instruction understanding, long-term context maintenance, and grounded trajectory generation, thereby translating complex language goals into temporally coherent and physically feasible flight behaviors.

% In addition, to evaluate the video reasoning capability of vision-language models on UAV FPV video data, FLIGHT introduces a \textbf{Pilot Reasoning task}. In this task, a VLM analyzes a UAV video ($V$) together with a textual instruction ($I$), infers the UAV's current state and its task plan for the next time interval, and outputs the analysis in natural language as (Analyse). Pilot Reasoning task can be formulated as a conditional video-language reasoning function:

% \begin{equation}
%   \pi_{\mathrm{vlm}} : (\mathrm{V}, I) \mapsto \mathrm{Analyse}_{\mathrm{text}} .
%   \label{eq:pilot_reasoning}
% \end{equation}

In addition, to evaluate the video reasoning capability of vision-language models (VLMs) on UAV FPV video data, FLIGHT introduces the \textbf{Pilot Reasoning task}. Given a UAV video ($V$) and a corresponding textual instruction ($I$), the VLM is required to infer the UAV's current state and its subsequent task plan. The model then generates this analysis in natural language, denoted as $\mathcal{T}_{\mathrm{reason}}$. Formally, the Pilot Reasoning task is formulated as a conditional text generation process:

\begin{equation}
\mathcal{T}_{\mathrm{reason}} = \pi_{\mathrm{vlm}}(V, I),
\label{eq:pilot_reasoning}
\end{equation}

where $\pi_{\mathrm{vlm}}$ represents the vision-language model that maps the multimodal inputs to the textual analysis space.

This task effectively evaluates the instruction understanding and planning capabilities of VLMs in UAV navigation scenarios, thereby providing a benchmark for assessing foundation models for navigation agents.

% where $\ell(\cdot, \cdot)$ is the task-specific loss function,
% $\Omega(\theta)$ is a regularization term, and $\lambda > 0$ controls
% the regularization strength.

\begin{figure}[!t]
  \centering
  \includegraphics[width=0.95\linewidth, keepaspectratio]{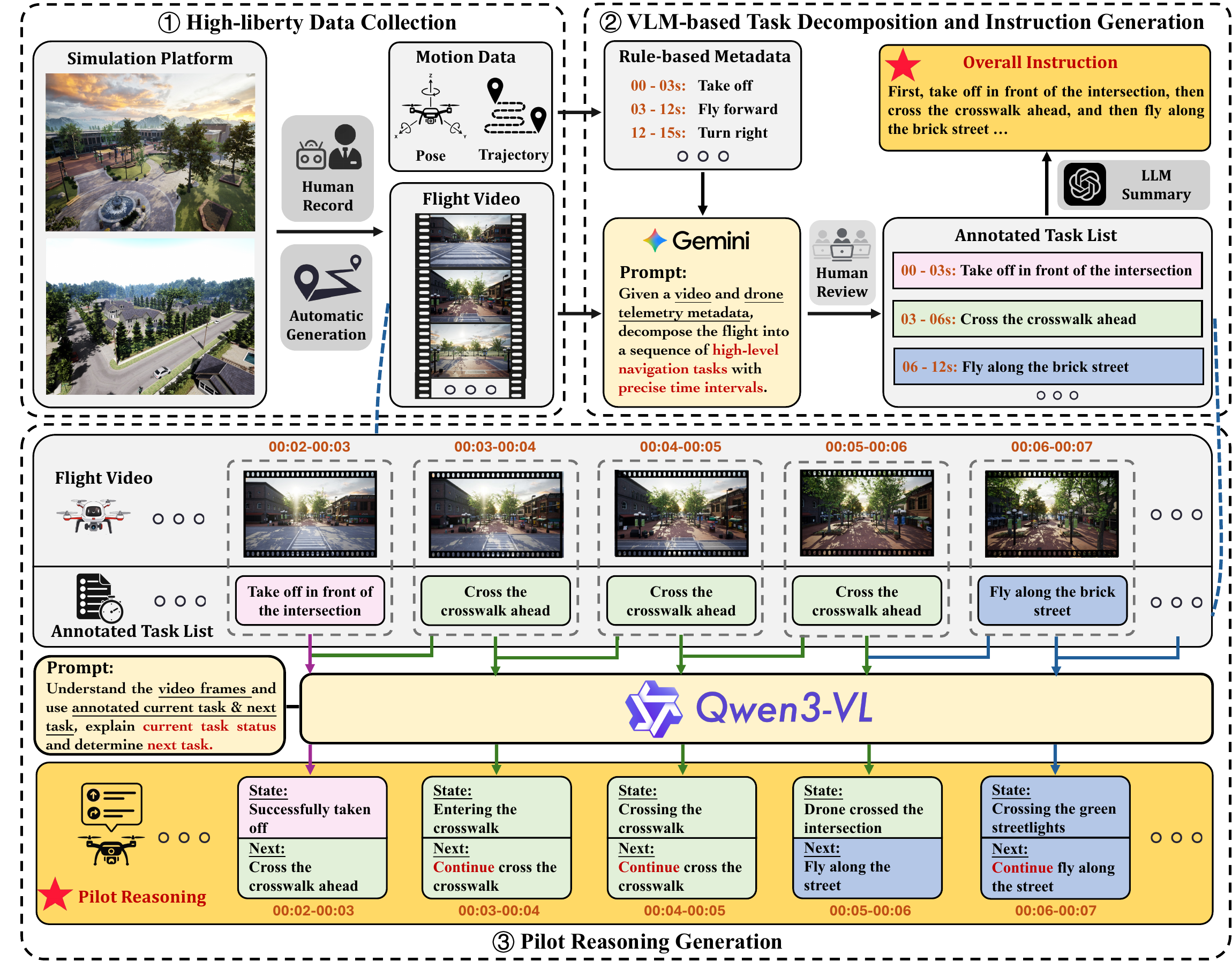}
  \caption{The Collection and Annotation pipeline of FLIGHT data. The pipeline consists of three core stages: (1) Trajectory data collection by human pilots  or automatic method within a simulation environment, followed by a rule-based extraction to generate metadata tags; (2) VLM-driven generation of task semantic labels for each video segment assisted by metadata tags, which are subsequently fused into task instruction texts; (3) Generation of Pilot Reasoning texts encompassing state analysis and look-ahead reasoning, achieved by combining video segments with the fused current and next-step task semantic labels.}
  \label{fig:annotation_pipeline}
\end{figure}

\subsection{Data Collection and Annotation}

\subsubsection{UAV Flight Trajectory and Video Data Collection}

For precise language-driven UAV navigation, we collect demonstration data across diverse, semantically rich simulation environments (urban, industrial, and rural) built via Unrealzoo ~\citep{zhong2024unrealzoo}. Featuring complex assets like realistic buildings and vegetation, these high-fidelity scenes provide the multi-modal variance necessary to effectively align textual instructions with grounded environmental semantics.

% The dataset covers diverse trajectories collected both by human pilots and automatically generated by path-planning models. 
The dataset covers diverse trajectories mainly collected by human pilots, with a small portion of simple-route scenarios supplemented by path-planning models. Unlike most UAV VLN benchmarks, such as Aerial VLN ~\citep{liu_2023_AerialVLN} and OpenFly~\citep{OpenFly}, which adopt discrete motion instructions, our dataset records full 6-DoF trajectory points, including coordinates (\text{latitude}, \text{longitude}, \text{altitude}) and orientation (\text{roll}, \text{pitch}, \text{yaw}). We further compute the relative displacement between consecutive trajectory points to obtain VLA-style action sequences.

Continuous motion commands enable more complex and flexible trajectory execution, resulting in smoother and more natural flight paths that better satisfy the requirements of fine-grained navigation. The simulator simultaneously records first-person-view (FPV) videos of UAV flights, which are temporally aligned with the trajectory-point data.

\subsubsection{Task-Semantic Segmentation Annotation and Instruction Generation}
In the data annotation stage, as illustrated in Step 2 of Figure \ref{fig:annotation_pipeline}, we aim to generate fine-grained task-oriented semantic descriptions for each segmented video clip. Specifically, each segment is annotated with both the drone’s motion behavior and the surrounding environmental context, such as "taking off from a plaza, crossing a pedestrian walkway, flying along a street past traffic signs, or turning left near a roadside kiosk". To this end, we feed the segmented video clips into a vision-language model (VLM) to produce segment-level annotations. This segment-level annotation strategy can more accurately capture fine-grained semantic changes throughout the flight process and is therefore better suited for fine-grained navigation and reasoning tasks.

Vision-only models often struggle to infer UAV attitude and motion trends from FPV videos. We therefore use UAV motion data as auxiliary prompts: raw signals such as attitude angles are analyzed to locate key action intervals, including takeoff, landing, and turning, and converted into metadata prompts alongside video clips for the VLM.

Compared with holistic video annotation and key-frame-based methods, our video-segmentation annotation pipeline with motion-prior assistance better preserves both fine-grained action details and the completeness of flight motion. Segment-level annotation helps capture subtle but critical navigation behaviors, while motion cues compensate for visual models’ limited spatial-motion understanding. This enables the pipeline to generate more complete and accurate fine-grained annotations for navigation tasks.

For the textual outputs of segment-level annotation, we use an LLM to merge multiple subtask descriptions into a complete high-level instruction. Rather than directly concatenating segment descriptions, the LLM rewrites them into coherent, natural instructions that better reflect human language. This encourages semantic-level instruction understanding over keyword matching or template-based reasoning.

\subsubsection{Construction of Pilot Reasoning Texts}
This module further generates the corresponding \textbf{Pilot Reasoning} texts based on the segment-level semantic annotations obtained in the previous stage. The core objective of Pilot Reasoning is to simulate task planning and decision-making during flight. Therefore, the generated text is expected not only to describe the UAV’s current flight state, but also to perform \textbf{prospective reasoning} about the actions that may be executed in the next time interval, thereby reflecting the capability of continuous task planning along the temporal dimension.

To achieve this goal, we introduce task-level prior information from an oracle future information during the Pilot Reasoning generation stage. As shown in the lower part of Fig. \ref{fig:annotation_pipeline}, the model receives both the task description of the current temporal segment, which serves as the textual representation of the current state, and the task description of the subsequent temporal segment, which provides prior information for future action planning. The vision-language model then combines these textual cues with the video clip content to jointly analyze the current flight state and future motion tendency, and generates a complete Pilot Reasoning text. The resulting reasoning output can therefore capture not only what the UAV is currently doing, but also what it is expected to do next, forming a temporally coherent and planning-oriented representation of flight decision-making.

\subsection{Data Analysis}

\begin{figure}[htbp]
  \centering
  \includegraphics[width=0.95\linewidth]{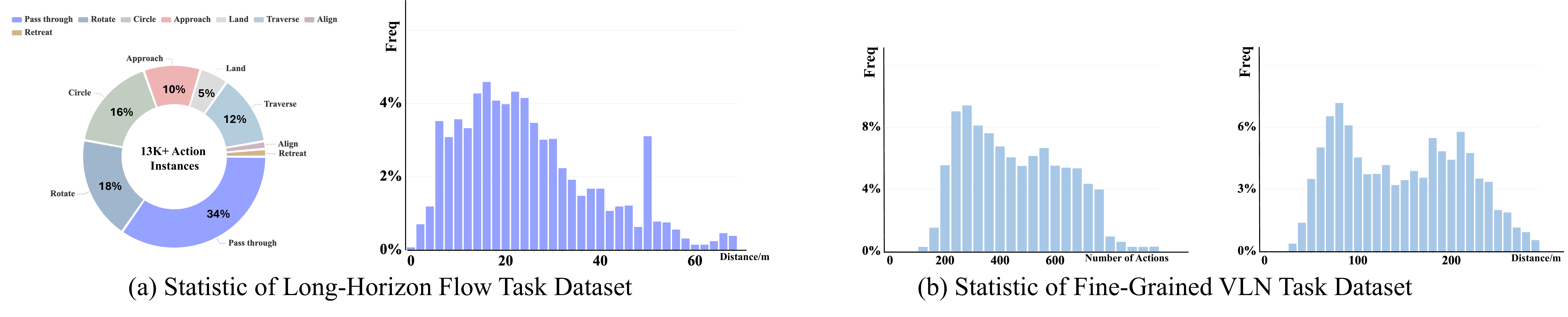}
  \caption{Dataset statistics for FLIGHT. (a) Long-Horizon Flow statistics, including sub-task categories and trajectory lengths. (b) Fine-Grained VLN trajectory profiles, including the number of actions per trajectory and total trajectory distance.}
  \label{fig:fgvln_data}
\end{figure}

% \begin{figure}[htbp]
%   \centering
%   \includegraphics[width=0.9\linewidth]{images/longflow.pdf}
%   \caption{Statistical characterization of the FLIGHT Long-Horizon Flow dataset. Left: Distribution of sub-task categories across the dataset. Right: Distribution of trajectory lengths in meters (m), for the Long-Horizon Flow tasks.}
%   \label{fig:long_flow}
% \end{figure}

We construct two trajectory datasets for long-horizon, fine-grained UAV navigation: \textbf{Fine-grained VLN} and \textbf{Long-horizon Flow}, containing 6,689 and 4,098 trajectories, respectively.

The Long-horizon Flow dataset inherits eight action categories from UAV-Flow \citep{nips2025_92cfa104}, with the category distribution shown in the left panel of Fig.~\ref{fig:fgvln_data}(a). It contains 13,815 action instances in total, with each trajectory averaging 3.37 action instances. The average destination distance is 17.8 m, and the average trajectory length is 32.8 m; their distributions are shown in the right panel of Fig.~\ref{fig:fgvln_data}(a). Some trajectories have near-zero displacement because the dataset includes rotation-centric actions, and algorithmically generated orbiting trajectories produce a few high-frequency displacement values. Overall, however, the unique trajectory lengths remain broadly distributed.

The Fine-grained VLN dataset features linguistically diverse instructions with dense spatial and action-level grounding. We sample continuous actions at 10 Hz, resulting in an average of 475 actions per trajectory and an average trajectory length of 154.5 m. Fig.~\ref{fig:fgvln_data}(b) reports the detailed distributions of action counts and trajectory lengths.

\section{FLIGHT VLA Pipeline}
\textbf{FLIGHT VLA} is an asynchronous VLA architecture for complex long-horizon UAV flight. Its key design goal is to retain long-term semantic context while meeting the high-frequency control demands of continuous flight. We therefore decompose the policy into two modules that operate at different temporal scales. A low-frequency \textbf{Streaming Pilot VLM} performs online video understanding, task-stage recognition, and high-level planning, while a high-frequency \textbf{Diffusion Action Model} generates continuous control trajectories from real-time visual observations and UAV states. The two modules communicate through asynchronous semantic feature transfer, allowing semantic reasoning to run at a lower rate than the control loop.

We begin by formulating the continuous control objective. Given a language instruction \(I\), FPV video frames \(\{V_t\}\), and IMU and pose states \(\{S_t\}\), the agent predicts a horizon of future actions at each control step \(t\):

\begin{equation}
  A_t 
  = \pi_\theta\!\left(I, V_{\leq t}, S_{\leq t}, M_{\leq t}\right) = \{a_t, a_{t+1}, \ldots, a_{t+H-1}\},
  \label{eq:streaming_full_policy}
\end{equation}

where \(H\) is the prediction horizon, each action \(a_t\) is a relative motion command in the local UAV coordinate frame, and \(\{M_t\}\) denotes the long-term contextual memory formed by historical reasoning texts. Directly using a large VLM to instantiate \(\pi_\theta\) at every control step is impractical because autoregressive reasoning introduces substantial latency. We instead factorize Eq.~\ref{eq:streaming_full_policy} into a slow semantic reasoning process and a fast action generation process:

\begin{align}
  (z_\tau, M_{\tau})
  &= g_\phi\!\left(I, V_{\tau-W_i:\tau}, S_{\tau-W_i:\tau}, M_{\tau-W_M:\tau}\right),
  \label{eq:streaming_pilot_vlm} \\
  A_t &= f_\theta\!\left(V_t, S_t, z_{\tau(t)}\right),
  \label{eq:streaming_action_model}
\end{align}

\begin{figure}[htbp]
  \centering
  \includegraphics[width=0.95\linewidth]{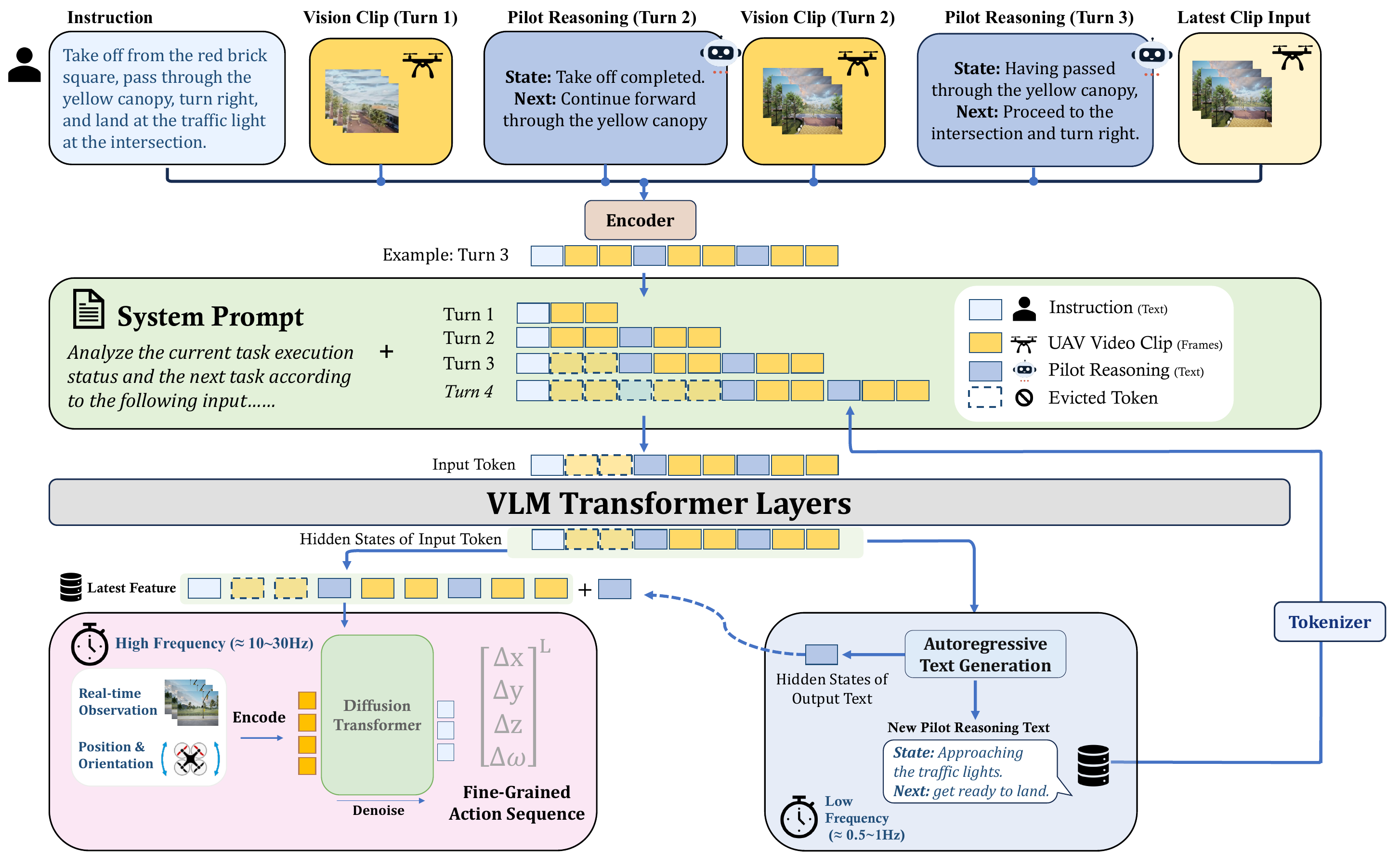}
  \caption{Overview of the proposed FLIGHT VLA framework. A low-frequency Streaming Pilot VLM continuously ingests multimodal inputs—including task instructions, historical memory, and visual frames—to perform online semantic reasoning and generate semantic features. Concurrently, a high-frequency Action Model fuses real-time visuals, UAV states, and the latest semantic features to generate fine-grained action trajectories via a DiT-based flow matching strategy. Through this asynchronous dual-rate architecture, the system decouples semantic planning from motion control, thereby balancing long-range task comprehension with real-time navigation capabilities.}
  \label{fig:streaming_framework}
\end{figure}

Here, \(g_\phi\) is the \textbf{Streaming Pilot VLM} and \(f_\theta\) is the \textbf{Action Model}. At reasoning time ($\tau$), the VLM consumes a recent video-state window of length ($W_i$) and a reasoning-memory window of length ($W_M$), and outputs a Pilot Reasoning text ($M_\tau$) along with a semantic latent feature ($z_\tau$), which encodes the hidden states of both the input and output reasoning texts, as illustrated in Fig.~\ref{fig:streaming_framework}.
 The function \(\tau(t)\) indexes the most recent VLM output available to the controller at time \(t\). This definition explicitly permits \(\tau(t) < t\), thereby modeling VLM inference latency while preserving high-frequency action generation.

\subsection{Streaming Pilot VLM for Online Task Reasoning}

\textbf{Streaming Pilot VLM} is the slow reasoning module of the system. Built upon StreamingVLM~\citep{xu2025streamingvlmrealtimeunderstandinginfinite}, an efficient online video understanding architecture, it maintains task context over a continuous UAV FPV stream and produces textual reasoning about the current flight state and the near-future plan. As shown in Eq.~\ref{eq:streaming_pilot_vlm}, each VLM update takes as input the task instruction \(I\), the latest video-state window \((V_{\tau-W_i:\tau}, S_{\tau-W_i:\tau})\), and the historical reasoning memory \(M_{\tau-W_M:\tau}\). This memory stores recent Pilot Reasoning texts, which compress long-horizon multimodal history into a compact textual representation. Consequently, \(W_M\) can be set relatively large, allowing the VLM to access a broader range of historical task memory without incurring the cost of retaining all past visual tokens.

The generated text ($M_\tau$) summarizes the current flight state and predicts the next action, such as subtask completion, turning, crossing an object, approaching a target, adjusting heading, hovering, or moving to the next subgoal. It is then interleaved with the corresponding visual frame ($V$) and current state ($S$), and written back to memory, as the format illustrated in the upper part of Fig.~\ref{fig:streaming_framework}. We also extract the final-layer hidden states associated with the model input and the most recent generated reasoning tokens, yielding the semantic feature \(z_\tau\). This hidden-state-to-action conditioning follows the general design of GR00T N1~\citep{gr00tn1_2025}, while our framework adapts it to asynchronous FLIGHT VLA control. This feature carries information about the instruction, historical context, current visual state, and next-step plan, and serves as the semantic condition for high-frequency action generation.

For training, we adopt an \textbf{overlapped-chunk full-attention} strategy. The long video stream is divided into consecutive temporally overlapping chunks. Within each chunk, historical Pilot Reasoning texts are concatenated with interleaved image-text sequences to construct an independent training instance, and full attention is applied over all tokens in the chunk. This preserves consistency between the training token layout and the inference-time streaming schedule. We supervise the VLM with Pilot Reasoning texts from the FLIGHT dataset, encouraging it to learn task-execution analysis and next-step action planning.

\subsection{Diffusion Action Model for Fine-Grained Continuous Control}
The \textbf{Action Model} serves as the fast execution module of the system and is responsible for generating continuous flight trajectories within the high-frequency control loop. At each control time step \(t\), this module receives three types of conditioning information: the current FPV frame \(V_t\), the UAV state \(S_t\), and the latest available semantic feature \(z_{\tau(t)}\) from the Streaming Pilot VLM. Here, \(z_{\tau(t)}\) is asynchronously updated by the slow VLM, while the image and state inputs are continuously refreshed at the control frequency. The timestamp \(\tau(t)\) denotes the most recent VLM decision output available at time \(t\) and satisfies \(\tau(t) < t\).

To provide high-frequency visual feedback, we equip the Action Model with an independent visual encoder rather than reusing the low-frequency visual features from the VLM. Specifically, the current image \(V_t\) is encoded by a pretrained Vision Transformer into a real-time visual feature:

\begin{equation}
  \mathbf{f}_t = E_{\mathrm{vit}}(V_t).
  \label{eq:visual_feature}
\end{equation}

This design enables the action model to continuously perceive local visual changes between two consecutive VLM reasoning steps, thereby rapidly correcting yaw deviations, target offsets, and dynamic viewpoint changes.

Meanwhile, to avoid drift in the global coordinate frame and allow the action model to adapt to VLM features produced at different historical time steps, we represent the current UAV state using a relative pose encoding. Specifically, the UAV pose at the current time step is expressed in the coordinate frame of the VLM reasoning time step, and the resulting relative pose is encoded into a vector representation $q_t$ for model processing. The final conditioning input to the Action Model is

\begin{equation}
  c_t = \left[\mathbf{f}_t;\, q_t;\, z_{\tau(t)}\right].
  \label{eq:action_condition}
\end{equation}

For action generation, we adopt a trajectory generator based on a \textbf{Diffusion Transformer}. Given the condition \(c_t\), the model outputs a continuous action sequence over the future horizon of \(H\) steps:

\begin{equation}
  A_t = f_\theta(c_t).
  \label{eq:conditioned_action_generation}
\end{equation}

Each action consists of relative displacement and attitude changes in the local coordinate frame. Since the model predicts an action sequence rather than a single-step control command, the system can continue to obtain valid control commands from the current action sequence even before the VLM completes its next round of reasoning, thereby avoiding control gaps caused by the inference latency of the model.

During training, the Action Model is supervised using continuous action segments from expert flight trajectories. We train the diffusion-based action generator with a \textbf{Flow Matching} objective, enabling the model to learn a conditional flow field from noisy trajectories to real trajectories. 

\subsection{Joint Training with Latency-Aware Alignment}
The training objective of \textbf{FLIGHT VLA} consists of both the VLM reasoning loss and the action generation loss:

\begin{equation}
  \mathcal{L}
  =
  \lambda_{\mathrm{vlm}}\mathcal{L}_{\mathrm{vlm}}
  +
  \lambda_{\mathrm{act}}\mathcal{L}_{\mathrm{act}},
  \label{eq:joint_training_objective}
\end{equation}

where \(\lambda_{\mathrm{vlm}}\) and \(\lambda_{\mathrm{act}}\) control the weights of the two supervision signals, respectively.

Because the pretrained VLM branch and the trajectory-trained Action Model converge at different rates, we use a two-stage training strategy. We first optimize both modules jointly so that the Action Model adapts to VLM semantic features, then freeze the stabilized VLM and continue training the Action Model to improve control accuracy while avoiding VLM overfitting.

\begin{figure}[htbp]
  \centering
  \includegraphics[width=0.95\linewidth]{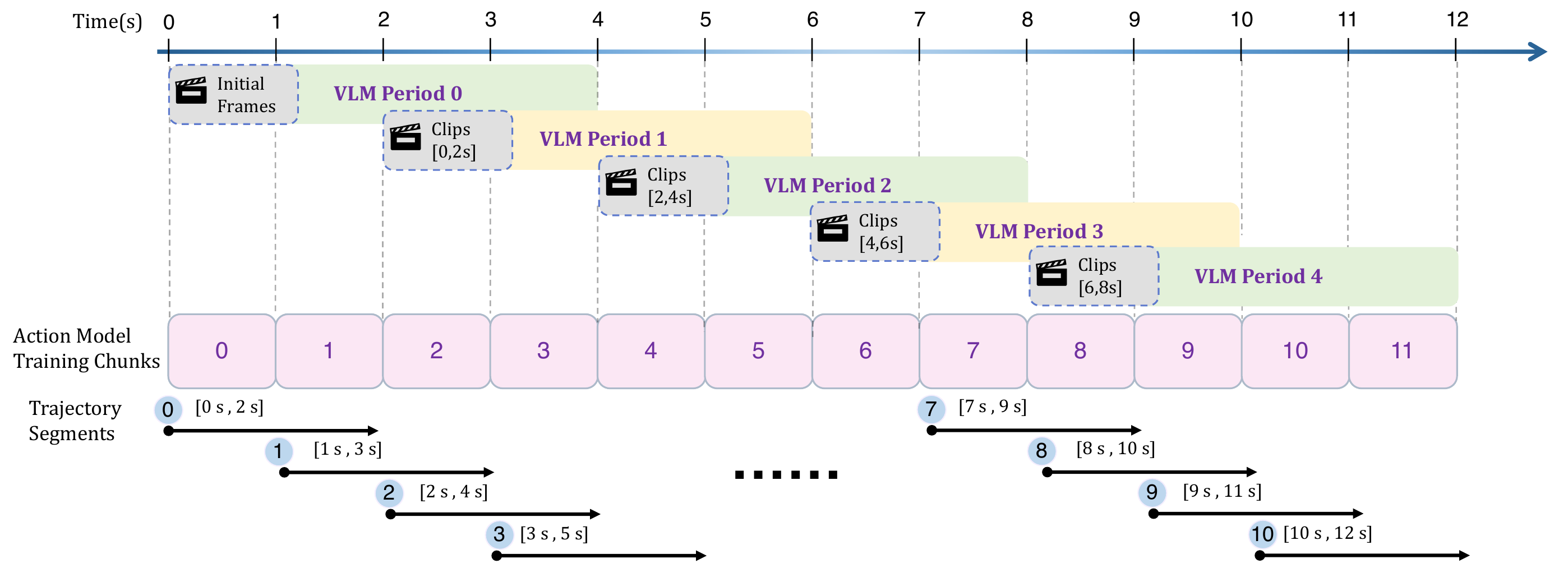}
  \caption{\textbf{Asynchronous Training Strategy for the FLIGHT VLA Framework.} We simulate asynchronous semantic feature arrival during training. The Streaming Pilot Reasoning VLM processes 2-second FPV history windows and produces low-frequency semantic features, while the action model is trained at a higher control frequency on dense trajectory chunks. For each action chunk, we use VLM features with different temporal offsets to mimic inference delays; e.g., at \(t=7\) s, the action model is supervised on the 7--9 s trajectory and conditioned on features from Period 3 with a 1-second delay or Period 2 with a 3-second delay.}
  \label{fig:ays_train}
\end{figure}

In real-world deployment, the inference latency of the VLM is not fixed. If training always assumes that the Action Model receives VLM features from the current time step, the model may suffer from distribution shift when exposed to stale semantic features during inference. To address this issue, we explicitly simulate asynchronous latency during training data construction. For each action training segment, we sample VLM features from several historical temporal offsets. As illustrated in Fig.~\ref{fig:ays_train}, VLM inputs are constructed using video windows with 1-second and 3-second delays for selected action-model training chunks, such as the 7th and 9th chunks, and 0-second and 2-second delays for the remaining chunks, excluding the initial and final chunks. The relative pose encoding is also computed using the associated VLM reasoning time as the local reference frame.

This training data organization strategy enables the Action Model to adaptively handle VLM semantic features originating from different historical time steps, and to learn to compensate for their temporal lag using current visual feedback and relative state information. Consequently, during real-world deployment, even when the VLM reasoning interval fluctuates, the Action Model can still stably generate control sequences that remain consistent with the current flight state.

%% ============================================================
%%  4. Experiments
%% ============================================================
\section{Experiments}
\subsection{Evaluation Metrics}

For the proposed \textbf{Long-horizon Flow} task, trajectory evaluation requires a more fine-grained assessment. A trajectory is expected not only to reach the final destination but also to accurately and completely execute each subtask specified by the instruction. The output action sequence must strictly conform to the language instruction. For example, if the instruction requires the UAV to traverse an object from the left side, the trajectory must indeed pass through the left side rather than merely completing a traversal action. The entire navigation task is counted as successful in SR only when all subtasks are completed as required. In addition, we report \textbf{Stage Success Rate (SSR)}, which measures the proportion of subtasks that are accurately completed within a trajectory.

We evaluate \textbf{Fine-Grained VLN} performance using standard metrics. Navigation Error (NE) measures the average distance between the agent's final position and the target. Success Rate (SR) measures the fraction of episodes in which the agent stops within 15 m of the target. Oracle Success Rate (OSR) considers an episode successful if the trajectory reaches within 15 m of the target at any point. Normalized Dynamic Time Warping (NDTW) measures the similarity between the predicted and reference trajectories, reflecting how well the agent follows the demonstrated path. Standard SR solely evaluate the final arrival status, ignoring whether the agent faithfully complies with the sequential constraints execution. To bridge this gap, we propose Instruction Adherence Success Rate (IASR) to strictly assess fine-grained execution fidelity. A navigation trial is deemed successful if and only if the agent sequentially fulfills all sub-tasks explicitly dictated in the instruction (e.g., takeoff, traversing specific obstacles, and landing).

We construct a navigation-oriented video QA benchmark by sampling UAV FPV videos of different lengths. Given a video from task start to the current step, the model have to infer the current navigation state and decide the next action. We use GPT-5 to assess the semantic consistency between the ground-truth and generated reasoning texts, and report the resulting accuracy as the model performance.

\begin{figure}[htbp]
  \centering
  \includegraphics[width=0.85\linewidth]{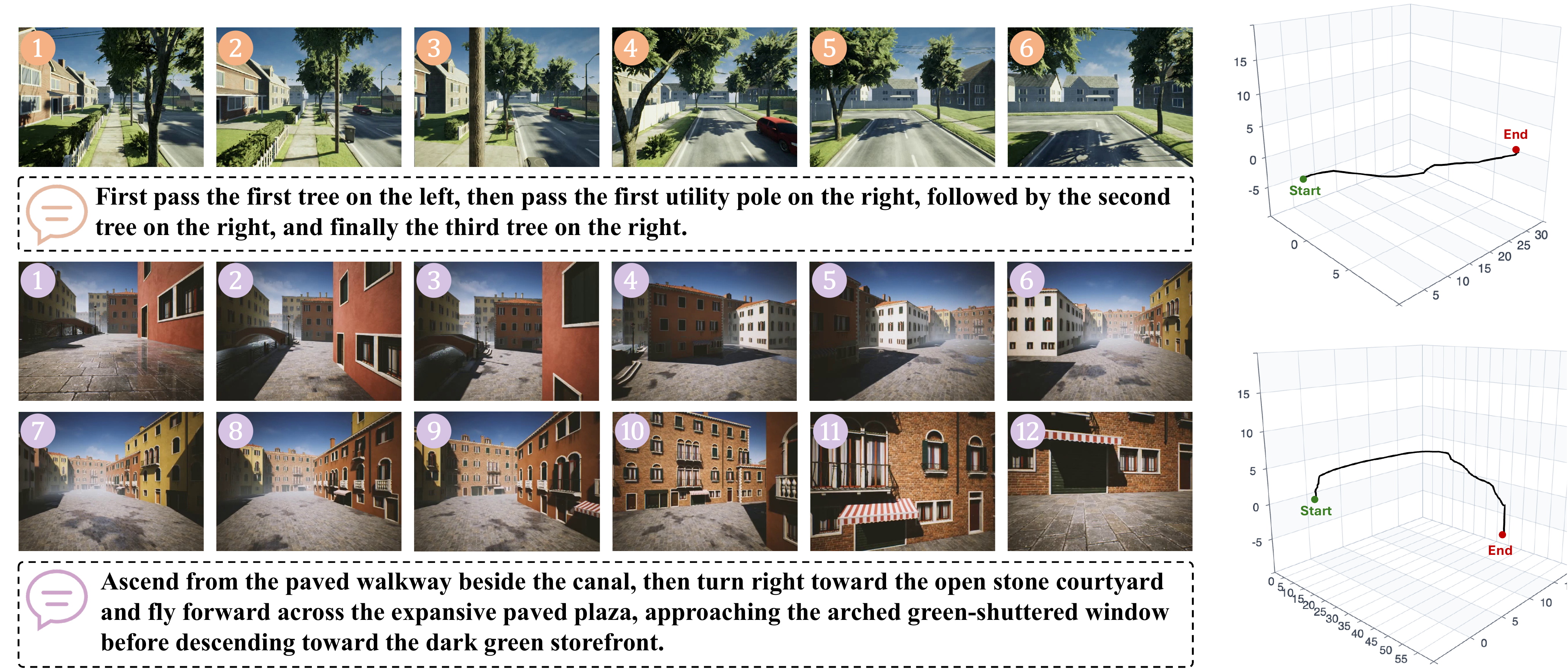}
  \caption{\textbf{Visualization results of our method.} First row demonstrate our UAV successfully follows the instruction in Fine-Grained Flow Task. Another two rows show the performance of our method in the Long-Horizon Flow task. The UAV can accurately perceive the execution status of each subtask and generate correct action sequences to complete the entire instruction.}
  \label{fig:results_visualization}
\end{figure}

\subsection{Benchmark Baseline}

We construct the UAV benchmark based on representative model paradigms from recent VLA and VLN literature. Given the differences between the tasks—VLN models are primarily designed for long-range navigation, whereas VLA models are mainly developed for ground robot manipulation—we structurally modify and reconfigure these models to meet the requirements of Long-Horizon Flow tasks and Fine-Grained VLN tasks.

\textbf{VLA Models.} We select following VLA base models for evaluating Long-Horizon Flow tasks.
(1) \textbf{OpenVLA} ~\citep{kim24openvla} is a prominent open-source 7B-parameter VLA model designed for generalizable robot manipulation. When adapting OpenVLA to our task, we retain its discrete token-based output format. Following the step-size range of UAV motion, we discretize the UAV motion space and map it to 256 tokens. We also encode the UAV state information and textual task instruction into a unified sentence, following the design of OpenVLA-UAV in UAV-Flow ~\citep{nips2025_92cfa104}.
(2) \textbf{Memory VLA} ~\citep{shi2025memoryvla} formulates robot manipulation in VLA models as a sequential decision-making process. This method introduces a memory architecture for storing historical information, making it more suitable for multi-stage long-horizon tasks. For adaptation, we use the current FPV image as the RGB input together with the language instruction. The output action space is defined as a 4-dimensional relative pose command. During inference, the model relies on its internal memory management mechanism to preserve historical information and support decision making.

\textbf{VLN Models.} We consider models originally designed for VLN tasks and adapt them to the Fine-Grained VLN setting. The selected baseline models are as follows.
(1) \textbf{CMA+LAG} ~\citep{liu_2023_AerialVLN} extends the Cross-Modal Attention (CMA) model~\citep{vln2018anderson}, which fuses language instructions with RGB-D observations for navigation decisions. Look-ahead Guidance (LAG) improves CMA by replacing shortest-path supervision to the final goal with guidance toward look-ahead target points constructed from the ground-truth trajectory.
(2) \textbf{NaVid} \citep{zhang2024navid} is the first video-based embodied large model specifically designed for VLN. It takes video observations collected during navigation and natural language instructions as input, directly generates reasoning text, and maps it to robot navigation actions.

\subsection{Result}
\textbf{Long-horizon Flow Task.} As shown in Table~\ref{tab:long_horizon_flow_results}, the proposed method outperforms the baseline models across all evaluation metrics. OpenVLA does not explicitly account for the execution requirements of long-horizon tasks in its architectural design and lacks an effective contextual memory mechanism. As a result, it struggles to continuously track the execution state of long-sequence tasks. Given visual observations, the model cannot reliably infer which subtask should be executed next within a long instruction sequence.

Although Memory VLA introduces a memory module to alleviate this issue, its memory mechanism remains insufficient for fully capturing historical information in semantically complex task scenarios. Consequently, its performance drops substantially when the instruction involves two or more subtasks. In addition, both baseline models have difficulty accurately identifying task termination points and cannot effectively determine whether all subtasks specified in the instruction have been completed.

In contrast, the proposed \textbf{FLIGHT VLA} architecture fully leverages the advantages of streaming video understanding models and supports an ultra-long context window. It successfully completes test instances containing up to seven subtasks. By continuously analyzing the FPV video stream, the vision-language model can accurately perceive and infer the execution status of each subtask. When the historical context indicates that all instructed tasks have been completed, the VLM module can generate a textual response to guide the action model to remain stationary. These results strongly demonstrate the feasibility and effectiveness of using online Video-LLM to guide UAVs in executing long-horizon, multi-stage tasks.

\textbf{Fine-grained VLN Task.} We then evaluate the models on the \textbf{Fine-grained VLN} task. Table~\ref{tab:fine_grained_vln_results} summarizes the navigation results. Compared with the Long-horizon Flow task, Fine-grained VLN involves substantially longer trajectories and is therefore more challenging. The CMA+LAG method is limited by its model capacity and exhibits insufficient understanding of complex instructions and multimodal inputs.

NaVid generates actions autoregressively with a VLM and relies only on historical images for context management. As a result, it suffers from a context-length bottleneck, and the model often begins to generate actions randomly after completing the first subtask. The model also struggles to perform precise landing at the instructed location, indicating insufficient fine-grained instruction understanding.

It is worth noting that both types of VLN baselines overlook deployment efficiency. At each step, the model predicts only a single discrete motion action instead of an action sequence; after executing it, the UAV must stop, wait for the next inference result, and then resume motion, making the protocol highly constrained by inference and transmission latency.

Driven by a streaming VLM~\citep{xu2025streamingvlmrealtimeunderstandinginfinite}, our method overcomes the context-length bottleneck in long-range navigation and is able to parse and decompose multi-stage tasks. Moreover, for fine-grained task instructions, our model can capture detailed textual constraints and accurately perform takeoff, navigation maneuvers, and precise landing. Unlike conventional baselines based on discrete control, our model uses continuous motion trajectories for navigation and balances real-time control efficiency through its slow-fast system design.

Fig.~\ref{fig:results_visualization} visualizes two navigation results of our method. Across both Long-horizon Flow and Fine-grained VLN tasks, the UAV follows the given instructions and completes the required navigation behaviors. In the first example, the UAV passes three objects from the specified directions and stops at the final target tree. In the second example, it completes the full instruction sequence, including takeoff, a right turn, forward flight, and landing. These behaviors are executed in the correct order and at the intended locations, demonstrating the model's ability to understand and perform complex instructions at a fine-grained level.

\begin{table}[htbp]
  \centering
  \caption{Comparison with baseline methods on two type of tasks}
  \label{tab:main_results}
  \begin{subtable}[t]{0.46\linewidth}
    \centering
    \caption{Results on Long-horizon Flow.}
    \label{tab:long_horizon_flow_results}
    \small
    \begin{tabular}{lccc}
      \toprule
      \textbf{Method} & \textbf{SR$\uparrow$ (\%)} & \textbf{SSR$\uparrow$ (\%)} & \textbf{NDTW$\uparrow$ (\%)} \\
      \midrule
      OpenVLA-UAV & 10.50 & 30.87 & 18.02 \\
      Memory VLA & 18.50 & 38.69 & 16.58 \\
      \textbf{FLIGHT VLA} & \textbf{59.00} & \textbf{78.97} & \textbf{45.78} \\
      \bottomrule
    \end{tabular}
  \end{subtable}
  \hfill
  \begin{subtable}[t]{0.52\linewidth}
    \centering
    \caption{Results on Fine-grained VLN task.}
    \label{tab:fine_grained_vln_results}
    \small
    \begin{tabular}{lcccc}
      \toprule
      \textbf{Method} & \textbf{SR$\uparrow$(\%)} & \textbf{OSR$\uparrow$(\%)} & \textbf{NE$\downarrow$(m)} & \textbf{IASR$\uparrow$(\%)} \\
      \midrule
      CMA+LAG & 6.5 & 22.0 & 52.56 & 4.5 \\
      NaVid & 13.0 & 25.5 & 46.69 & 5.0\\
      \textbf{FLIGHT VLA} & \textbf{13.5} & \textbf{37.0} & \textbf{45.52} & \textbf{11.0} \\
      \bottomrule
    \end{tabular}
  \end{subtable}
\end{table}

\textbf{VLM Accuracy.} We evaluate representative VLMs on VLN video-text sequences, as these videos are typically longer and contain richer instructional and visual semantics, thereby providing a more suitable benchmark for assessing video models’ long-range semantic understanding and planning capabilities. As reported in Table~\ref{tab:vlm_reasoning_accuracy}, after training with the streaming reasoning paradigm, the models exhibit substantially improved understanding and reasoning capabilities on navigation data. The task-level semantic guidance provided by the VLM lays the foundation for the action model to generate motion trajectories that are consistent with the given instructions.

\begin{table}[htbp]
  \centering
  \caption{Reasoning accuracy of representative VLMs on VLN video-text sequences.}
  \label{tab:vlm_reasoning_accuracy}
  \begin{tabular}{lccc}
    \toprule
    \textbf{Model} & \textbf{Parameters} & \textbf{State Acc.(\%)} & \textbf{Action Acc.(\%)} \\
    \midrule
    Gemini-3-flash & -- & 25.8 & 39.1 \\
    Qwen 3.6 plus & -- & 2.9 & 10.6 \\
    Qwen 2.5 VL & 3B & 5.0 & 6.9 \\
    \textbf{Streaming Pilot Reasoning VLM} & 3B & \textbf{64.2} & \textbf{68.1} \\
    \bottomrule
  \end{tabular}
\end{table}

\section{Conclusion}
In this work, we introduced FLIGHT, a comprehensive benchmark for long-horizon, fine-grained, instruction-guided UAV navigation and reasoning tasks. FLIGHT addresses the limitations of prior UAV benchmarks by integrating continuous fine-grained action sequences with rich language instructions, encompassing both long-horizon flow tasks and fine-grained visual-language navigation (VLN) tasks. To support robust model training and evaluation, we also introduced Pilot Reasoning, a textual reasoning signal derived from human pilot trajectories, providing explicit supervision for task planning and decision-making.
We further proposed the FLIGHT VLA architecture, which combines a streaming visual-language reasoning backbone with a diffusion-based action model, organized under a fast-slow asynchronous system. This design enables real-time, continuous control while preserving long-range semantic understanding and reasoning capabilities. Experimental results across both Long-Horizon Flow and Fine-Grained VLN tasks demonstrate that our approach outperforms baseline VLA and VLN models, validating the efficacy of streaming reasoning and the integration of explicit pilot supervision.

{\small
\bibliographystyle{\colabbibstyle}
\bibliography{references}
}

%% ============================================================
%%  Appendix  (Requirement 6 — tcolorbox demo)
%% ============================================================
\appendix
% \section{Additional Experimental Details}

% \subsection{Hyperparameter Sensitivity}

% We study the sensitivity of our method to key hyperparameters. The temperature
% parameter $\tau$ in the contrastive loss is set to 0.07 based on a grid search
% over $\{0.03, 0.05, 0.07, 0.10, 0.15\}$. The regularization coefficient
% $\lambda$ is set to $10^{-4}$ following standard practice.

% \subsection{Dataset Statistics}

% \begin{table}[htbp]
%   \centering
%   \caption{Statistics of the five evaluation benchmarks.}
%   \label{tab:datasets}
%   \begin{tabular}{lrrrl}
%     \toprule
%     \textbf{Dataset} & \textbf{Train} & \textbf{Val} & \textbf{Test} & \textbf{Task} \\
%     \midrule
%     Dataset-A & 50,000  & 5,000  & 10,000 & Classification \\
%     Dataset-B & 120,000 & 12,000 & 24,000 & Detection \\
%     Dataset-C & 30,000  & 3,000  & 6,000  & Segmentation \\
%     Dataset-D & 80,000  & 8,000  & 16,000 & Retrieval \\
%     Dataset-E & 200,000 & 20,000 & 40,000 & Generation \\
%     \bottomrule
%   \end{tabular}
% \end{table}

% %% ============================================================
% %%  Appendix B — tcolorbox Environments Demo (Requirement 6)
% %% ============================================================
\section{Prompt Engineering in Data Annotation and Pilot Reasoning VLM design}

% \begin{colabinsight}
% The core insight behind our approach is that hierarchical representations
% combined with contrastive pre-training produce features that are both
% discriminative and transferable across tasks.  The multi-scale fusion
% module is the critical component enabling this synergy.
% \end{colabinsight}

% \begin{colabtakeaway}
% When deploying representation learning models, practitioners should:
% \begin{enumerate}[leftmargin=1.5em]
%   \item Pre-train with contrastive objectives on large unlabeled corpora.
%   \item Use hierarchical features instead of single-scale extraction.
%   \item Fine-tune end-to-end with the combined loss for best results.
% \end{enumerate}
% \end{colabtakeaway}

% \begin{colablimitations}
% \begin{itemize}
%   \item The pre-training phase requires 8$\times$A100 GPUs for 72 hours,
%         which may be prohibitive for smaller labs.
%   \item Performance gains diminish on datasets with fewer than 1,000 labeled
%         samples.
%   \item The method has not been tested on modalities beyond vision and language.
% \end{itemize}
% \end{colablimitations}

\begin{colabprompt}[Task Decomposition Prompt]
    
    You are an expert drone flight instructor. Your task is to generate a precise, step-by-step flight log based on the provided sequence of timestamped frames extracted from the FPV flight and the provided Flight Telemetry Data.

    \#\#\# 1. APPROVED ACTION VOCABULARY (STRICT) \\
    **You must ONLY use action verbs from the following list to describe the drone's movements.**

    * **Basic Movement:** Fly forward, Move forward, Head towards, Advance, Proceed straight, Go straight, Keep straight, Maintain heading, Continue, Keep flying, Resume, Ascend, Rise, Climb, Gain altitude, Take off, Descend, Lower, Drop down, Lose altitude, Land.\\
    * **Turning (Use ONLY for significant directional changes):** Turn right, Turn left, Make a right, Make a left, Yaw right, Yaw left, Bear right, Bear left.
    * **Complex Maneuvers (Telemetry Only):** Perform a U-turn, Circle once, Circle twice, Orbit, Ascend, Descend.\\
    * **Orientation:** Follow, Trace, Fly parallel to, Approach, Head for, Close in on, Align with, Face, Orient towards, Line up with.\\
    * **Interaction:** Pass, Pass by, Go past, Fly past, Leave behind, Pass between, Go through the gap of, Thread through, Continue past, Overtake, Go beyond, Fly over, Pass over, Soar over, Clear, Fly through, Traverse, Enter, Fly under, Pass beneath, Go under.\\

    \#\#\# 2. MANDATORY TELEMETRY INJECTION (HIGHEST PRIORITY)\\
    The following complex maneuvers were physically recorded by sensors. The logs now contain two time ranges (in seconds): the **ACTION TIME** (where the turn or vertical move occurred) and the **CONTEXT** time (which includes the action time plus \{context seconds\} seconds before and after, for visual reference).

    **CRITICAL INSTRUCTION:** You MUST visually locate these moments and insert the action **EXACTLY** as written in the logs. **You MUST use the wider CONTEXT time range to identify the surrounding landmarks and add a contextual description of the location to the action.**

    {turn telemetry str}

    **CRITICAL RULES FOR TELEMETRY:**\\
    1.  **NO SIMPLIFICATION/MODIFICATION:** Use the action phrase **EXACTLY** as written (e.g., "Circle once Right", "Ascend while Turning Left").\\
    2.  **MUST ADD LOCATION:** The final step output must combine the action and a precise description of the surroundings (e.g., "perform a U-turn Left at the edge of the blue-roofed market.").\\
    3.  **TIMESTAMP REQUIREMENT:** You MUST append the specific `[ACTION TIME]` from the log to the end of the step in the format `[start s - end s]`.\\
    4.  **LOCATION EXAMPLE (REQUIRED FORMAT):**
        * Log: `- [ACTION TIME: 4.00s - 6.00s] [CONTEXT: 2.00s - 8.00s] MANDATORY ACTION: perform a U-turn Left.`
        * Output: `perform a U-turn Left near the tiered stone fountain. [4.00s-6.00s]`\\

    \#\#\# 3. MOVEMENT SENSITIVITY \& FILTERING\\
    * **IGNORE MINOR ADJUSTMENTS:** Do **NOT** report small yaw corrections, camera jitters, or slight steering (< 30 degrees). Treat these as **"Fly forward"**.\\
    * **MAJOR VISUAL TURNS:** Only report a visual turn NOT in the logs if it is a **major, obvious turn**.\\

    \#\#\# 4. OBJECT DIVERSITY \& SALIENCY\\
    - **Prioritize Prominent Landmarks:** Focus on the most visually distinct objects (e.g., large buildings, unique statues, colorful signs).\\
    - **Avoid Repetition:** Shift focus to **different** nearby objects in consecutive steps.\\
    - **Dynamic Attention:** Point out new features in each step rather than mentioning the same object repeatedly.\\

    \#\#\# 5. STRICT SENTENCE STRUCTURE \& TIMING\\
    Except for the mandatory telemetry actions, every step must strictly follow this pattern:\\
    **[Action Verb] + [Preposition] + [Visual Description] + [Object] [Start\_Time s - End\_Time s]**

    * **TIMING RULE (OVERLAP ALLOWED):** Actions often happen simultaneously. **It is permitted for timestamps to overlap.** For example, Step 2 may start before Step 1 ends if the actions are concurrent.
    * **Correct Examples:**\\
        * "Fly over the white fountain. [1.2s-3.5s]"\\
        * "Pass between the blue market stalls. [3.0s-6.0s]" (Note: Starts before previous step ends)\\
        * "Turn right towards the brick building. [6.0s-8.2s]" (Only if major turn)\\

    \#\#\# 6. NEGATIVE CONSTRAINTS (FORBIDDEN)\\
    - **NO TEXT RECOGNITION:** Do NOT output specific text characters (e.g., "UFoods", "Stop"). Describe visually (e.g., "the red brick building").\\
    - **NO Bold Text**: Do not use asterisks (**).\\
    - **NO Meta-Comments**: Start **directly** with "1.".\\
    - **NO Degree Numbers**: Use qualitative terms for angles.\\

    \#\#\# 7. OUTPUT FORMAT EXAMPLE\\
    1. Ascend from the paved area. [0.00s-3.50s]\\
    2. Fly forward over the circular fountain. [2.50s-5.10s]\\
    3. Proceed straight through the open plaza. [4.80s-8.00s] \\
    4. Pass between the covered stalls. [4.60s-10.20s]\\
    5. Approach the large tree. [10.00s-12.50s]\\
    6. perform a U-turn Left near the tiered stone fountain. [12.50s-15.00s]\\
    7. Circle once Right above the market area. [12.00s-19.00s] \\
    8. Fly along the pedestrian street. [18.50s-22.10s]
    ...

    Generate the Numbered List now:
    
\end{colabprompt}

\begin{colabprompt}[Instruction Annotation Prompt]
    
    You are an expert drone flight narrator.

    \#\#\# INPUT DATA\\
    1. **Visuals**: The sequence of timestamped frames provided.\\
    2. **Step-by-Step Flight Log**:
    \{step\_list\_text\}

    \#\#\# TASK\\
    Convert the Step-by-Step Flight Log into a **Single, Coherent Global Mission Instruction** (one continuous narrative).

    \#\#\# STRICT REQUIREMENTS\\
    1. **Preserve All Details**: You MUST include **EVERY** object and **EVERY** descriptive adjective mentioned in the flight log. Do not generalize or omit details (e.g., "blue and white striped awnings" must NOT be shortened to "awnings").\\
    2. **Handle Simultaneity (Critical)**: Watch the video carefully to identify actions from the list that happen at the same time.\\
        - *Example*: If the drone flies under tree branches and passes a lamp post at the same moment, connect them using **"while"**, **"as"**, or **"simultaneously"** (e.g., "Pass under the green tree branches **while** passing by the black lamp post...").\\
    3. **Flow**: Connect all steps into a fluent, natural narrative using varied transition words.\\
    4. **Constraint**: Do not add new information not present in the list, but do not lose any information from the list.\\
    5. **Timestamp Removal**: Do NOT include the timestamps [0.0s-0.0s] in this narrative summary. Keep it as pure text.

    \#\#\# OUTPUT FORMAT \\
    Output ONLY the global instruction text.
    
\end{colabprompt}

\begin{colabprompt}[Pilot Reasoning Annotation Prompt]
 You are a professional drone flight assistant. Analyze the following inputs: Previous Modal inference results(History),three sequentially sampled frames from the video, Overall Mission List, Current Mission, and Human Pilot's Decision for next mission.
    Assume no knowledge of the human pilot's reasoning or declared next decision. Based on the visual context from the video, provide a single, concise paragraph analyzing why the pilot chooses the next mission. Your analysis must ONLY address:
    Completion Status: Has the current mission been completed based on the visual evidence?
   \\ \#\#\# ACTION PLAN:\\
    - If YES: How to execute the next step based on the pilot's decision.\\
    - If NO: How to proceed to finish the current mission.\\
    Constraint: Direct output only. Do not include introductory or concluding remarks.

    \#\#\# OUTPUT RULES:\\
    - Keep your output strictly under 80 English words. Output English only.\\
    - Do not include the word "pilot's" in the output.\\
    - Output **only** the reasoning process. Ground all reasoning strictly in observable visual context; do not assume intent or external knowledge beyond what is shown.\\
    - Concise and clear.\\
    - No commentary, no additional formatting\\
    - NO precise timestamps like 00:06.53 or 7.33s\\
    - DO NOT include task numbers in the output.
\end{colabprompt}

\begin{colabprompt}[Pilot Reasoning VLM Prompt]
You are a professional drone flight assistant. Analyze the following inputs: Previous Modal inference results(History), the video frames, overall mission instruction. Based on the visual context from the video, provide a single, concise paragraph analyzing which action or mission should be executed next. **Constraints:**  1. **Format:** \\`Status: [State]. Next: [Action].\\` 2. **Length:** Total output must be **under 30 words**. 3.Direct output only. Do not include introductory or concluding remarks. The overall mission instruction is: \{Instruction\}. Following is the modal inference history: /n \{History with frames, states and previous Pilot Reasoning Text\}
\end{colabprompt}

\begin{colabprompt}[Critic GPT Prompt]
            \textbf{State score prompt}:
            "You are evaluating whether a predicted instruction sentence matches the ground-truth "
            "instruction sentence semantically."
            "Focus on whether the key meaning is consistent, including:"
            "- important objects, locations, direction, actions and mission progress"
            "Treat paraphrases or minor wording differences as equivalent."
            "Treat contradictions, missing crucial actions, wrong objects/locations, or clearly "
            "different mission state as not equivalent."
            "Return JSON only with this exact schema:"
            \'\{"equivalent": true/false, "reason": "short explanation", "confidence": 0.0\}\'
            f"Ground-truth Status:\{gt text\}"
            f"Predicted Status:{pred text}"
            \newline

            \textbf{Action score prompt:}
            "You are evaluating whether a predicted instruction sentence matches the ground-truth "
            "instruction sentence semantically."
            "Focus on whether the key meaning is consistent, including:"
            "- important objects, locations, direction, actions and mission progress"
            "Treat paraphrases or minor wording differences as equivalent."
            "Treat contradictions, missing crucial actions, wrong objects/locations, or clearly "
            "different mission state as not equivalent."
            "Return JSON only with this exact schema:"
            '\{"equivalent": true/false, "reason": "short explanation", "confidence": 0.0\}'
            f"Ground-truth Status:\{gt text\}"
            f"Predicted Status:\{pred text\}"
        )
\end{colabprompt}

\section{Examples instructions of FLIGHT Task}
\label{instruction_example}

This section provides more examples of the task definied in FLIGHT task.

\begin{table}[htbp]
  \centering
  \caption{Comparison between conventional VLN instructions and the
           fine-grained VLN instructions in FLIGHT.}
  \label{tab:vln_instruction_comparison}
  \small
  \setlength{\tabcolsep}{4pt}
  \renewcommand{\arraystretch}{1.12}
  \begin{tabularx}{\linewidth}{@{}P{0.17\linewidth}Y P{0.27\linewidth}@{}}
    \toprule
    \textbf{Source} & \textbf{Instruction example} & \textbf{Task property} \\
    \midrule
    Conventional VLN &
    Start along the street, turn left at the intersection, fly across the bridge,
    and finally land. &
    Coarse route-level navigation with sparse grounding cues. \\
    \addlinespace[2pt]
    Fine-grained VLN &
    Start along the street, pass the blue umbrella, turn left at the intersection,
    fly across the bridge, and finally land beside the second tree on the left. &
    Fine-grained navigation with dense landmark grounding and precise terminal
    localization. \\
    \bottomrule
  \end{tabularx}
\end{table}

\begin{table}[htbp]
  \centering
  \caption{Comparison between single-stage UAV-Flow instructions and the
           multi-stage compositional instructions in FLIGHT.}
  \label{tab:instruction_comparison}
  \small
  \setlength{\tabcolsep}{4pt}
  \renewcommand{\arraystretch}{1.12}
  \begin{tabularx}{\linewidth}{@{}P{0.39\linewidth}Y@{}}
    \toprule
    \textbf{UAV-Flow} & \textbf{Long-horizon Flow} \\
    \midrule
    Circle around the person in a clockwise direction. &
    (1) First circle around the person in a clockwise direction, and then circle a
    second target in a different direction. \\
    & (2) Circle around the person \textbf{three times} clockwise.\\
    \addlinespace[2pt]
    Approach the person from the left side &
    Pass the tree in front from the right, then approach the person from the left
    side, and finally land.
     \\
     \addlinespace[2pt]
     Pass through the tree from the right side & Pass through the first tree on the right, then pass the second tree on the left, and finally land beside the third tree.
     \\
    \bottomrule
  \end{tabularx}
\end{table}

\FloatBarrier

\section{Training Details}
\label{sec:training_details}

This section provides the complete training configuration for the training of the FLIGHT VLA model.

\subsection{Model Architecture}

\textbf{Base Vision-Language Model.} We initialize our model from Qwen2.5-VL-3B-Instruct and train it with low-rank adaptation (LoRA).

\textbf{Action Prediction Head.} We adopt a DiT-B (Diffusion Transformer) based action head. The action head configuration is summarized in Table~\ref{tab:action_head_config}.

\begin{table}[htbp]
  \centering
  \caption{Action head configuration.}
  \label{tab:action_head_config}
  \small
  \setlength{\tabcolsep}{8pt}
  \renewcommand{\arraystretch}{1.12}
  \begin{tabularx}{0.64\linewidth}{@{}Y P{0.34\linewidth}@{}}
    \toprule
    \textbf{Parameter} & \textbf{Value} \\
    \midrule
    Action dimension & 4 \\
    State dimension & 4 \\
    Hidden dimension & 1024 \\
    Future action window size & 7 \\
    Repeated diffusion steps & 2 \\
    \bottomrule
  \end{tabularx}
\end{table}

\subsection{Training Configuration}

During training, videos are sampled at 2 FPS. The visual window lengths for Long-Horizon Flow and Fine-Grained VLN are set to 12 seconds and 8 seconds per training chunk, respectively. For text processing, we employ 512 text sink tokens alongside a sliding window of 512 tokens. To balance the optimization rates—since the VLM converges faster than the DiT action model—the joint co-training loss is defined as:
\begin{equation}
  \mathcal{L}
  =
  0.1\mathcal{L}_{\mathrm{vlm}}
  +
  \mathcal{L}_{\mathrm{act}},
  \label{eq:joint_training_objective}
\end{equation}

Table~\ref{tab:training_hyperparams_lhf}, ~\ref{tab:training_hyperparams_vln} summarize the detailed hyperparameters used for training FLIGHT VLA on the Long-Horizon Flow and Fine-Grained VLN datasets.

\begin{table}[htbp]
  \centering
  \caption{Training hyperparameters of Long Horizon Flow.}
  \label{tab:training_hyperparams_lhf}
  \small
  \setlength{\tabcolsep}{8pt}
  \renewcommand{\arraystretch}{1.12}
  \begin{tabularx}{0.78\linewidth}{@{}Y P{0.42\linewidth}@{}}
    \toprule
    \textbf{Hyperparameter} & \textbf{Value} \\
    \midrule
    Learning rate & $5 \times 10^{-5}$ \\
    Batch size per device & 8 \\
    Gradient accumulation steps & 2 \\
    Effective batch size & 128 (8 $\times$ 2 $\times$ 8 GPUs) \\
    GPU & $8 \times$ A800 80G \\
    \bottomrule
  \end{tabularx}
\end{table}

\begin{table}[htbp]
  \centering
  \caption{Training hyperparameters of Fine-Grained VLN.}
  \label{tab:training_hyperparams_vln}
  \small
  \setlength{\tabcolsep}{8pt}
  \renewcommand{\arraystretch}{1.12}
  \begin{tabularx}{0.78\linewidth}{@{}Y P{0.42\linewidth}@{}}
    \toprule
    \textbf{Hyperparameter} & \textbf{Value} \\
    \midrule
    Learning rate & $1 \times 10^{-4}$ \\
    Batch size per device & 8 \\
    Gradient accumulation steps & 2 \\
    Effective batch size & 128 (8 $\times$ 2 $\times$ 8 GPUs) \\
    GPU & $8 \times$ A100 80G \\
    \bottomrule
  \end{tabularx}
\end{table}

\FloatBarrier

\end{document}